\pdfoutput=1

\documentclass[11pt]{article}

\usepackage{EMNLP2023}

\usepackage{times}
\usepackage{latexsym}
\usepackage{bm}

\usepackage[T1]{fontenc}

\usepackage[utf8]{inputenc}

\usepackage{microtype}

\usepackage{inconsolata}

\usepackage{xspace}
\usepackage{booktabs}
\usepackage{amsmath}
\usepackage{amssymb}
\usepackage{graphicx}
\usepackage{adjustbox}
\usepackage{multirow}
\usepackage{comment}
\usepackage{color, colortbl}
\usepackage{arydshln}
\usepackage{hhline}
\usepackage{array}
\usepackage{makecell}

\DeclareMathOperator*{\argmax}{arg\,max}

\newcommand{\rparagraph}[1]{\vspace{1.2mm}\noindent\textbf{#1.}}
\newcommand{\sparagraph}[1]{\vspace{0.0mm}\noindent\textbf{#1.}}

\newcommand{\zsxlt}{{\textsc{zs-xlt}}\xspace}

\newcommand{\xlt}{{\textsc{xlt}}\xspace}

\newcommand{\last}{{\textsc{last}}\xspace}
\newcommand{\srcdev}{{\textsc{src-dev}}\xspace}
\newcommand{\trgdev}{{\textsc{trg-dev}}\xspace}

\newcommand{\sdev}{{\textsc{s-dev}}\xspace}

\newcommand{\ca}{{\textsc{ca}}\xspace}

\title{One For All \& All For One: Bypassing Hyperparameter Tuning with Model Averaging For Cross-Lingual Transfer}

\author{Fabian David Schmidt\textsuperscript{1}, Ivan Vulić\textsuperscript{2}, Goran Glavaš\textsuperscript{1} \\
  \textsuperscript{1} Center For Artificial Intelligence and Data Science, University of Würzburg, Germany \\
  \textsuperscript{2} Language Technology Lab, University of Cambridge, UK \\
  \texttt{\{fabian.schmidt, goran.glavas\}@uni-wuerzburg.de} \\
  \texttt{iv250@cam.ac.uk} }

\begin{document}
\maketitle
\begin{abstract}
Multilingual language models enable zero-shot cross-lingual transfer (\zsxlt): fine-tuned on sizable source-language task data, they perform the task in target languages without labeled instances.
The effectiveness of \zsxlt hinges on the linguistic proximity between languages and the amount of pretraining data for a language. Because of this, model selection based on source-language validation is unreliable: it picks model snapshots with suboptimal target-language performance.  
As a remedy, some work optimizes \zsxlt by extensively tuning hyperparameters: the  
follow-up work then routinely struggles to replicate the original results.
Other work searches over narrower hyperparameter grids, reporting substantially lower performance. In this work, we therefore propose an unsupervised evaluation protocol for \zsxlt that decouples performance maximization from hyperparameter tuning. As a robust and more transparent alternative to extensive hyperparameter tuning, we propose to \textit{accumulatively average} snapshots from different runs into a single model.
We run broad \zsxlt experiments on both higher-level semantic tasks (NLI, extractive QA) and a lower-level token classification task (NER) and find that conventional model selection based on source-language validation quickly plateaus to suboptimal \zsxlt performance. On the other hand, our accumulative run-by-run averaging of models trained with different hyperparameters boosts \zsxlt performance and closely correlates with ``oracle'' \zsxlt, i.e., model selection based on target-language validation performance.
\end{abstract}

\section{Introduction and Motivation}
\label{sec:intro}
Massively multilingual transformers (MMTs) like XLM-\{R,V\} \citep{conneau-etal-xlmr, fair-xlmv} or mT5 \citep{xue-etal-2021-mt5} are pretrained via language modeling on vast corpora encompassing 100+ languages.
MMT fine-tuned on labeled task data in a source language can transfer cross-lingually \textit{zero-shot}, i.e. without further annotations, to target languages \citep{hu2020xtreme, lauscher-etal-2020-zero}. However, pretraining corpora size and linguistic distance between the source and target language dictate the quality of \xlt \citep{lauscher-etal-2020-zero}.
This is why model selection based on source-language validation data unreliably correlates with \zsxlt and selects checkpoints that yield suboptimal target-language performance \citep{keung-etal-2020-dont}.
Worse yet, there is no ``best practice'' for replicating \zsxlt results of prior work.
Some works, as our results suggest (cf. \S\ref{sec:results-and-discussion}), (1) exhaust extraordinarily large hyperparameter grids and (2) monitor target-language performance for the best transfer (i.e., violating ``true'' \zsxlt) to outperform baselines \citep{conneau-etal-xlmr, hictl}.
Other works rerun baselines with little to no hyperparameter tuning \citep{hu2020xtreme,wu-dredze-2020-explicit}: the re-evaluation then often trails original results by non-negligible margins.\footnote{For instance, when evaluating XLM-V$_\text{base}$, \citet{fair-xlmv} have been unable to reproduce the original results of XLM-R$_\text{base}$ on the XNLI benchmark \citep{conneau-etal-xlmr}.} 
%
As a remedy, \citet{keung-etal-2020-dont} propose to evaluate \zsxlt on the snapshot that generalizes best to validation data in the target language (``oracle'' \zsxlt): as such, oracle \zsxlt stabilizes evaluation and denotes ideal transfer performance.
Nonetheless, 
oracle \zsxlt overstates the performance of true \zsxlt, for which no target-language instances are available \citep{schmidt2023free}. If they are, target-language annotations are always better levered for training than for validation \citep{schmidt-etal-2022-dont}.

This calls for an evaluation protocol that (1) maximizes ``true'' \zsxlt results and (2) makes them easily reproducible, regardless of the extent of hyperparameter tuning. 
%
%
In this work, we find that model averaging fulfills both criteria.
%
Weights averaging has proven effective in, e.g., MT \citep{vaswani2017nips} and recently NLU \citep{wang-etal-2022-adamix,schmidt2023free}.
\citet{schmidt2023free} enable model averaging for sizable gains in \xlt.
They first fine-tune an MMT on labeled source-language data
and then re-train models (i.e., more runs) by copying and freezing 
the task head of the initially fine-tuned model: this aligns snapshots and enables weight averaging across runs.\footnote{Fine-tuning models with different randomly initialized task heads otherwise yields sets of incompatible weights, hindering meaningful model averaging.}

\rparagraph{Contributions} 
In this work, we propose an evaluation protocol that decouples maximizing \zsxlt performance from hyperparameter tuning. The key idea is to \textit{accumulatively average} snapshots of runs with different hyperparameters: this improves performance over model selection based on source-language validation performance. We run exhaustive experiments on higher-level (NLI, extractive QA) and lower-level (NER) NLU tasks on a broad grid of hyperparameters and show, examining the cross-section of all runs, that model selection based on source-language validation almost exclusively picks snapshots suboptimal for \zsxlt. We also confirm that conventional hyperparameter tuning on source-language validation prematurely settles for models that maximize source-language performance at the expense of \zsxlt. Crucially, we show that accumulative model averaging performs on par or better than the best snapshot picked by source-language validation already from the second (i.e. first averaged-in) run and then consistently improves \zsxlt with more runs. We additionally show that this accumulative model averaging closely correlates with oracle \zsxlt \textit{without} requiring any source- or target-language validation data to maximize transfer performance. 

\section{Accumulative Run Averaging}
\label{sec:accumulative-averaging}
Prior work conducts model selection for \zsxlt by extensive hyperparameter tuning using either source- or target-language validation data. Whereas the latter violates true \zsxlt \cite{schmidt-etal-2022-dont}, the former overfits to source-language performance \cite{keung-etal-2020-dont}.  
The recent success of snapshot averaging in \xlt \citep{schmidt2023free} motivates our research question: can 
(accumulative) averaging of models trained during hyperparameter search outperform -- with fewer overall training runs -- the \zsxlt performance of the ``optimal'' model selected based on source-language validation performance?

We benchmark model selection based on source-language validation against accumulative model averaging as follows. We iteratively sample models (i.e., runs) $\{\{\theta_1, \ldots, \theta_{r}\} \; | \;  1 \leq r \leq 10\}$ with different hyperparameters (i.e., pairs of learning rates and batch sizes) from the pool of runs containing $N$ runs per hyperparameter configuration (cf.\,Appendix \S\ref{appendix:full-results}). We repeat this procedure $10$ times and report mean \zsxlt performance. 
Standard model selection picks the model $\{\argmax_{i}{{\text{Val}(\theta_i)}}  \; | \; 1 \leq i \leq r\}$ at run $r$ that maximizes source- (target-) language validation, capturing the ``true'' (``oracle'') \zsxlt performance.
``Accumulative averaging'', in contrast, naively averages (i.e. without any supervision) all models of $r$ runs to a single model $\frac{1}{r} \sum_{j=1}^{r}{\theta_j}=\bar{\theta}_r$. 

\section{Experimental Setup}
\label{sec:experimental-setup}

\sparagraph{Tasks and Languages}
We select for our evaluation two higher-level semantic tasks (NLI and and extractive QA) and one lower-level structured prediction task (NER). For each task, we fine-tune the MMT on the provided English training splits.\footnote{Train portion of MNLI~\citep{williams-etal-2018-broad}, the enclosed 3,696 English training instances of TyDiQA-GoldP for QA, and the English training portion of WikiANN for NER.}

\vspace{1mm}
\noindent \textit{Natural Language Inference} (NLI).
We evaluate NLI on XNLI~\citep{conneau2018xnli} and IndicXNLI~\citep{https://doi.org/10.48550/arxiv.2204.08776} which together cover 25 typologically diverse languages. 

\vspace{1mm}
\noindent \textit{Extractive QA} (TyDiQA-GoldP).
TyDiQA-GoldP comprises questions that are answered by a span of text in the provided gold passage and covers 9 diverse languages \citep{clark-etal-2020-tydi}.

\vspace{1mm}
\noindent \textit{Named Entity Recognition} (NER).
We evaluate NER on 25 languages from WikiANN~\citep{pan-etal-2017-cross}, 10 languages from MasakhaNER~\citep{adelani-etal-2021-masakhaner}, and  9 languages from MasakhaNER 2.0 \citep{adelani-etal-2022-masakhaner}.

\rparagraph{Training Details}
We train XLM-R$_{\text{large}}$ \citep{conneau-etal-xlmr} for 10 epochs with AdamW~\citep{adamw}, weight decay of $0.01$, gradient norm clipping to $1.0$, and a LR schedule of 10\% linear warm-up and decay.\footnote{The training data of TyDiQA-GoldP consists of merely 3,696 instances; we thus fine-tune longer, for 40 epochs.}
We save 10 snapshots per model, one at every $10\%$ of total training steps. The maximum sequence length is 128 tokens for NLI and NER and 384 with a stride of 128 for TyDiQA-GoldP.

\rparagraph{Hyperparameter Grids}
We simulate conventional hyperparameter grid search over a broad set of 21 configurations, pairing seven learning rates $l \in \{0.1, 0.5, 1, 1.5, 2, 2.5, 3\}e^{{-}5}$ with three batch sizes $b \in \{16, 32, 64\}$. The grid is deliberately kept wide and same for all tasks to not reflect any prior knowledge on task-specific ``good values''.\footnote{Our full results in Table \ref{tab:xlt-full-results} indicate that for each task we obtain maximal (oracle) \zsxlt performance with a different, task-specific hyperparameter configurations.} We retrain MMT for each pair $(l, b)$ three times with different random seeds to account for variances over individual runs.



\rparagraph{Model Variants}
We evaluate four model variants: $v \in $ \{\last, \srcdev, \ca, \textsc{trg-dev}\}.
\last is simply the final snapshot of a training run.
\srcdev is the snapshot that maximizes source-language validation performance~\citep{hu2020xtreme}.
\ca averages all snapshots of a run to a single model and, according to \citet{schmidt2023free}, outperforms \last and \srcdev.
\trgdev breaches ``true'' \zsxlt and picks the snapshot that performs best on the target-language validation data~\citep{keung-etal-2020-dont}: as such, it generally represents an upper-bound of single-run \zsxlt performance.\footnote{Irrespective of tasks and language, labeled instances in the target-language bring larger gains if used for training rather than for model selection~\citep{schmidt-etal-2022-dont}.}


\section{Results and Discussion}
\label{sec:results-and-discussion}
\begin{table*}[t!]
 \footnotesize
  \begin{adjustbox}{width=\linewidth,center}
   \setlength{\tabcolsep}{0pt}
     \begin{tabular}{c|ccc:ccc|ccc:ccc|ccc:ccc}
      \toprule
\multicolumn{1}{c}{}  & \multicolumn{6}{c}{\textbf{NLI}} & \multicolumn{6}{c}{\textbf{TyDiQA-GoldP}} & \multicolumn{6}{c}{\textbf{NER}} \\
\cmidrule(lr){2-7} \cmidrule(lr){8-13} \cmidrule(lr){14-19}

 \multicolumn{1}{c}{}  &
\multicolumn{3}{c}{\textbf{Max. \srcdev}} & \multicolumn{3}{c}{\textbf{Accumulative Averaging}} &
\multicolumn{3}{c}{\textbf{Max. \srcdev}} & \multicolumn{3}{c}{\textbf{Accumulative Averaging}} &
\multicolumn{3}{c}{\textbf{Max. \srcdev}} & \multicolumn{3}{c}{\textbf{Accumulative Averaging}} \\
\cmidrule(lr){2-4} \cmidrule(lr){5-7}
\cmidrule(lr){8-10} \cmidrule(lr){11-13}
\cmidrule(lr){14-16} \cmidrule(lr){17-19}

$\bm{r}$ &  
\textbf{\textsc{last}} & \textbf{\sdev} & \textbf{\ca} &
\textbf{\textsc{last}} & \textbf{\sdev} & \textbf{\ca} &
\textbf{\textsc{last}} & \textbf{\sdev} & \textbf{\ca} &
\textbf{\textsc{last}} & \textbf{\sdev} & \textbf{\ca} &
\textbf{\textsc{last}} & \textbf{\sdev} & \textbf{\ca} &
\textbf{\textsc{last}} & \textbf{\sdev} & \textbf{\ca} \\ \hline

\rule{0pt}{10pt}    $1$ & $76.5_{0.6}$ & $76.5_{0.8}$ & $77.3_{0.4}$ & $76.5_{0.6}$ & $76.5_{0.8}$ & $77.3_{0.4}$ & $71.9_{0.4}$ & $71.9_{0.7}$ & $73.6_{1.9}$ & $71.9_{0.4}$ & $71.9_{0.7}$ & $73.6_{1.9}$ 
&  $\mathbf{40.8_{2.7}}$ & $41.1_{3.1}$          & $\mathbf{44.6_{2.1}}$ & $40.8_{2.7}$                                & $41.1_{3.0}$                                & $44.6_{2.1}$                                \\  
\rule{0pt}{10pt}    $2$ & $77.2_{0.3}$ & $77.5_{0.4}$ & $\mathbf{77.6_{0.2}}$ & \cellcolor{green!15}$77.6_{0.3}$ & \cellcolor{green!45}{$77.8_{0.4}$} & \cellcolor{green!45}{$78.0_{0.2}$} & $71.9_{0.6}$ & $71.6_{0.6}$ & $73.3_{2.0}$ & $73.4_{1.2}$ & $73.3_{1.1}$ & $72.9_{2.8}$
&  $39.3_{2.1}$          & $39.3_{2.1}$          & $43.5_{1.1}$          & $43.2_{2.2}$                                & $43.2_{2.2}$                                & \cellcolor{green!45}{$45.6_{1.5}$}          \\  
\rule{0pt}{10pt}    $3$ & $77.2_{0.3}$ & $77.5_{0.4}$ & $77.6_{0.2}$ & \cellcolor{green!45}{{$77.8_{0.3}$}} & \cellcolor{green!45}{$77.9_{0.4}$} & \cellcolor{green!45}{$78.1_{0.2}$} & $72.1_{0.8}$ & $71.8_{0.8}$ & $74.1_{1.0}$ & $74.1_{0.7}$ & \cellcolor{green!15}{$74.2_{0.7}$} & $73.8_{1.2}$
&  $39.3_{1.2}$          & $39.5_{1.7}$          & $44.0_{1.1}$          & \cellcolor{green!45}{$45.0_{1.7}$}          & \cellcolor{green!45}{$45.1_{1.8}$}          & \cellcolor{green!45}{$47.3_{1.3}$}          \\  
\rule{0pt}{10pt}    $4$ & $77.2_{0.4}$ & $77.5_{0.4}$ & $77.5_{0.2}$ & \cellcolor{green!45}{$77.7_{0.3}$} & \cellcolor{green!45}{$77.9_{0.4}$} & \cellcolor{green!45}{$78.1_{0.3}$} & $72.5_{0.8}$ & $\mathbf{72.0_{0.9}}$ & $73.9_{0.6}$ & \cellcolor{green!45}{$74.5_{0.6}$} & $74.1_{0.4}$ & $74.1_{0.9}$ 
&  $40.2_{2.0}$          & $40.8_{2.2}$          & $44.5_{1.5}$          & \cellcolor{green!45}{$45.0_{1.7}$}          & \cellcolor{green!45}{$45.3_{1.8}$}          & \cellcolor{green!45}{$47.2_{1.4}$}          \\  
\rule{0pt}{10pt}    $5$ & $77.3_{0.3}$ & $\mathbf{77.6_{0.3}}$ & $77.5_{0.2}$ & \cellcolor{green!45}{$77.9_{0.2}$} & \cellcolor{green!45}{$78.0_{0.2}$} & \cellcolor{green!45}{$78.1_{0.1}$} & $\mathbf{72.6_{0.8}}$ & $72.0_{0.9}$ & $73.8_{0.8}$ & \cellcolor{green!45}{\underline{$\mathbf{74.7_{0.7}}$}} & \cellcolor{green!15}{$\mathbf{74.4_{0.6}}$} & \cellcolor{green!15}{$74.2_{0.8}$} 
&  $40.3_{2.0}$          & $\mathbf{41.2_{2.3}}$ & $43.9_{1.9}$          & \cellcolor{green!45}{$45.3_{1.7}$}          & \cellcolor{green!45}{$45.5_{1.7}$}          & \cellcolor{green!45}{$47.5_{1.4}$}          \\  
\rule{0pt}{10pt}    $6$ & $77.3_{0.3}$ & $77.6_{0.1}$ & $77.5_{0.2}$ & \cellcolor{green!45}{$77.9_{0.1}$} & \cellcolor{green!45}{$78.0_{0.2}$} & \cellcolor{green!45}{$78.1_{0.1}$} & $72.6_{0.8}$ & $72.0_{0.9}$ & $74.2_{0.5}$ & \cellcolor{green!45}{$74.7_{0.7}$} & \cellcolor{green!15}{$74.4_{0.5}$} & \cellcolor{green!15}{$74.2_{0.7}$}
&  $40.3_{2.0}$          & $41.2_{2.3}$          & $43.9_{1.9}$          & \cellcolor{green!45}{$45.7_{1.4}$}          & \cellcolor{green!45}{$45.9_{1.4}$}          & \cellcolor{green!45}{$47.9_{1.2}$}          \\  
\rule{0pt}{10pt}    $7$ & $77.3_{0.3}$ & $77.6_{0.1}$ & $77.5_{0.2}$ & \cellcolor{green!45}{$77.9_{0.2}$} & \cellcolor{green!45}{$78.1_{0.2}$} & \cellcolor{green!45}{$78.2_{0.2}$} & $72.3_{0.9}$ & $71.7_{0.7}$ & $\mathbf{74.3_{0.3}}$ & \cellcolor{green!45}{$74.6_{0.7}$} & \cellcolor{green!15}{$74.3_{0.6}$} & \cellcolor{green!15}{$74.2_{0.5}$} 
&  $40.0_{2.1}$          & $40.6_{2.3}$          & $44.1_{2.0}$          & \cellcolor{green!45}{$46.0_{1.3}$}          & \cellcolor{green!45}{$46.1_{1.3}$}          & \cellcolor{green!45}{$48.1_{1.1}$}          \\  
\rule{0pt}{10pt}    $8$ & $77.3_{0.3}$ & $77.6_{0.2}$ & $77.5_{0.2}$ & \cellcolor{green!45}{$\mathbf{78.0_{0.2}}$} & \cellcolor{green!45}{$\mathbf{78.2_{0.1}}$} & \cellcolor{green!45}{$\mathbf{\underline{78.3_{0.2}}}$} & $72.1_{0.9}$ & $71.7_{0.7}$ & $74.2_{0.5}$ & \cellcolor{green!45}{$74.6_{0.7}$} & \cellcolor{green!15}{$74.3_{0.5}$} & \cellcolor{green!15}{$\mathbf{74.3_{0.5}}$}
&  $40.0_{2.1}$          & $40.6_{2.3}$          & $44.6_{1.7}$          & \cellcolor{green!45}{$46.0_{1.1}$}          & \cellcolor{green!45}{$46.1_{1.2}$}          & \cellcolor{green!45}{$48.2_{1.0}$}          \\  
\rule{0pt}{10pt}    $9$ & $\mathbf{77.4}_{0.2}$ & $77.6_{0.2}$ & $77.6_{0.2}$ & \cellcolor{green!45}{$78.0_{0.1}$} & \cellcolor{green!45}{$78.1_{0.1}$} & \cellcolor{green!45}{$78.3_{0.2}$} & $72.0_{1.1}$ & $71.7_{0.7}$ & $74.2_{0.5}$ & \cellcolor{green!45}{$74.6_{0.5}$} & \cellcolor{green!15}{$74.4_{0.4}$} & \cellcolor{green!15}{$74.2_{0.4}$}
&  $39.6_{2.3}$          & $39.9_{2.4}$          & $44.3_{1.8}$          & \cellcolor{green!45}{$46.0_{0.6}$}          & \cellcolor{green!45}{$46.1_{0.7}$}          & \cellcolor{green!45}{$48.3_{0.7}$}          \\  
\rule{0pt}{10pt}   $10$ & $77.3_{0.2}$ & $77.6_{0.2}$ & $77.6_{0.2}$ & \cellcolor{green!45}{$78.0_{0.1}$} & \cellcolor{green!45}{$78.2_{0.1}$} & \cellcolor{green!45}{$78.3_{0.1}$} & $72.0_{1.1}$ & $71.7_{0.7}$ & $74.2_{0.5}$ & \cellcolor{green!45}{$74.6_{0.6}$} & \cellcolor{green!15}{$74.4_{0.4}$} & \cellcolor{green!15}{$74.2_{0.6}$} &  $39.6_{2.3}$          & $39.9_{2.4}$          & $44.4_{1.7}$          & \cellcolor{green!45}{$\mathbf{46.1_{0.5}}$} & \cellcolor{green!45}{$\mathbf{46.2_{0.6}}$}& \cellcolor{green!45}{\underline{$\mathbf{48.4_{0.5}}$}} \\  

      \bottomrule
    \end{tabular}
  \end{adjustbox}
  \caption{
  $\{\{\theta_1, \ldots, \theta_{r}\} \; | \; 1 \leq r \leq 10\}$ models sampled for variants $v \in $ \{\last, \srcdev, \textsc{ca}\} from Table \ref{tab:xlt-full-results} (cf. \S\ref{sec:experimental-setup}).
  ``Max. \srcdev'' picks the run $\{\argmax_{i}{{\text{SrcVal}(\theta_i^v)}}  \; | \; 1 \leq i \leq r\}$.
  ``Accumulative averaging'' simply averages all runs $\frac{1}{r} \sum_{j=1}^{r}{\theta_{j}^{v}}$.
  Metrics: accuracy for NLI, span-$F_1$ for TyDiQA and token-level $F_1$ for NER. Subscripts denote std. deviation.
  Colored averaging \colorbox{green!45}{outperforms +0.2 or more} or \colorbox{green!15}{performs $\pm$0.1 of} the best Max.\,\srcdev model.
  }
  \label{tab:xlt-averaging}
  \vspace{-0.5em}
\end{table*}

\begin{table}[t!]
  \begin{adjustbox}{width=\columnwidth,center}
    \setlength{\tabcolsep}{2pt}
    \begin{tabular}{c|cc:cc|cc:cc|cc:cc}
      \toprule
      \multicolumn{1}{c}{} & \multicolumn{4}{c}{\textbf{NLI}} &\multicolumn{4}{c}{\textbf{TyDiQA-GoldP}} & \multicolumn{4}{c}{\textbf{NER}} \\
\cmidrule(lr){2-5} \cmidrule(lr){6-9} \cmidrule(lr){10-13}
\multicolumn{1}{c}{} & \multicolumn{2}{c}{\textbf{Max. \textsc{dev}}} & \multicolumn{2}{c}{\textbf{Acc. Avg.}} & \multicolumn{2}{c}{\textbf{Max. Dev}} & \multicolumn{2}{c}{\textbf{Acc. Avg.}} & \multicolumn{2}{c}{\textbf{Max. \textsc{dev}}} & \multicolumn{2}{c}{\textbf{Acc. Avg.}} \\  
\cmidrule(lr){2-3} \cmidrule(lr){4-5} \cmidrule(lr){6-7} \cmidrule(lr){8-9} \cmidrule(lr){10-11} \cmidrule(lr){12-13}
      \textbf{ } & \textbf{\textsc{src}} & \textbf{\textsc{trg}} & \textbf{} & \textbf{\textsc{}} & \textbf{\textsc{src}} & \textbf{\textsc{trg}} & \textbf{} & \textbf{\textsc{}} & \textbf{\textsc{src}} & \textbf{\textsc{trg}} & \textbf{} & \textbf{\textsc{}}  \\
      $\bm{r}$ & \textbf{\textsc{dev}} & \textbf{\textsc{dev}} & \textbf{\ca} & \textbf{\textsc{soup}} & \textbf{\textsc{dev}} & \textbf{\textsc{dev}} & \textbf{\ca} & \textbf{\textsc{soup}} & \textbf{\textsc{dev}} & \textbf{\textsc{dev}} & \textbf{\ca} & \textbf{\textsc{soup}}  \\\hline

\rule{0pt}{12pt}      1 & $77.3$          &  $77.0$          & $77.3$          & $76.8$          & $71.9$          & $72.8$          & $73.6$          & $73.7$          & $41.1$          & $46.5$          & $44.6$          & $42.3$ \\
\rule{0pt}{12pt}      3 & $77.5$          &  $77.7$          & $78.1$          & $77.6$          & $71.8$          & $73.5$          & $73.8$          & $73.8$          & $39.5$          & $49.2$          & $47.3$          & $42.1$ \\
\rule{0pt}{12pt}      5 & $\mathbf{77.6}$ &  $77.9$          & $78.1$          & $77.6$          & $\mathbf{72.0}$ & $73.4$          & $\mathbf{74.2}$ & $\mathbf{74.3}$ & $\mathbf{41.2}$ & $49.7$          & $47.5$          & $\mathbf{42.8}$ \\
\rule{0pt}{12pt}      7 & $77.6$          &  $78.2$          & $78.2$          & $\mathbf{77.8}$ & $71.7$          & $73.7$          & $74.2$          & $73.9$          & $40.6$          & $\mathbf{49.9}$ & $48.1$          & $42.8$ \\
\rule{0pt}{12pt}     10 & $77.6$          &  $\mathbf{78.4}$ & $\mathbf{78.3}$ & $77.7$          & $71.7$          & $\mathbf{73.9}$ & $74.2$          & $73.8$          & $39.9$          & $49.9$          & $\mathbf{48.4}$ & $42.8$ \\
      
      \bottomrule
    \end{tabular}
  \end{adjustbox}
  \caption{
  ``Max.\,\trgdev'' selects the run $\{\argmax_{i}{{\frac{1}{|T|}\text{Val}_{T}(\theta_i)}}  \; | \; 1 \leq i \leq r\}$, where $T$ is the set of target languages.
  \textsc{soup} averages the five checkpoints (from all available runs) that ``Max.\,\srcdev''.
  For other details, see Table \ref{tab:xlt-averaging}.
  }
  \label{tab:xlt-validation}
  \vspace{-0.8em}
\end{table}


\sparagraph{Single-Run Performance} The full \zsxlt results by hyperparameters are presented in Appendix \S\ref{appendix:full-results} (cf. Table \ref{tab:xlt-full-results}).
We observe that optimal \zsxlt of single runs depends on all axes of analysis: task, hyperparameters, and model variant.
While \last and \srcdev generally perform well, their \zsxlt performance fluctuates substantially across hyperparameter configurations, in line with \citep{keung-etal-2020-dont, schmidt2023free}.
\ca is a strong and robust baseline that often outperforms \last and \srcdev by notable margins on TyDiQA and NER. In the context of a single run, \ca performs especially well with suboptimal hyperparameters, even sometimes outperforming \trgdev.\
We also confirm that \ca remedies variation in \zsxlt both within and across hyperparameters \citep{schmidt2023free}. 
Table \ref{tab:xlt-full-results} (cf. Appendix \S\ref{appendix:full-results}) further highlights the notable gap in \zsxlt performance between the best-performing hyperparameter configurations and those selected based on source-language validation.
Only the ``oracle'' model selection based on target-language validation reliably correlates with the actual best (test) \zsxlt performance. 

\rparagraph{Run-by-Run Analysis} Table \ref{tab:xlt-averaging} compares \zsxlt performance run-for-run of all variants for model selection based on source-language validation (Max.\,\srcdev) against our accumulative averaging of randomly sampled runs with different hyperparameters (cf. \S\ref{sec:accumulative-averaging}).
%
On NLI, picking a single model on source-language validation only improves \zsxlt when moving from having one to having two models (i.e., between first two rows of Table \ref{tab:xlt-averaging}) and stagnates when having more models to choose from. 
With more runs, source-language validation may even prefer models that are worse at \zsxlt on TyDiQA and NER. Conventional model selection thus maximizes source-language performance at the expense of \zsxlt.
%
Across the board, accumulative averaging already matches or surpasses Max.\,\srcdev (with any number of models) using merely two or three runs. 
Moreover, accumulative averaging consistently outperforms the \textit{overall best} single-run model chosen from $3+$ runs (highlighted in green), irrespective of the task.
On all tasks, accumulative averaging stabilizes \zsxlt and reduces performance variance vis-a-vis Max.\,\srcdev counterparts.

Accumulatively averaging within-run snapshots (\ca) outperforms \last and \srcdev slightly on NLI and materially on NER. 
For NER, \zsxlt from WikiANN to MasakhaNER (2.0) also represents a domain transfer (from Wiki to news), in which \ca yields tremendous gains. In-domain (i.e., test on WikiANN), \ca generally performs on par with \last and \srcdev. 
The same is not true for QA, where \ca performs slightly worse: we ascribe this to averaging of ``unconverged'' snapshots, owing to the small TyDiQA training set (merely 3,696 instances), especially from runs with smaller learning rates and larger batches (cf.\,Table \ref{tab:xlt-full-results}).

\rparagraph{Further Analyses} 
Table \ref{tab:xlt-validation} extends the run-by-run analysis to \trgdev and ``model soups'' (\textsc{soup}) to illustrate why accumulative model averaging outperforms model selection based on source-language validation. Rather than selecting a single snapshot, \textsc{soup} averages the five snapshots (among all available runs) with best  source-language validation performance \citep{pmlr-v162-wortsman22a}.

Compared to (oracle) \trgdev, accumulatively averaging runs performs on par on NLI, slightly better on TyDiQA, and somewhat worse on NER.
\trgdev selects language-specific snapshots, thereby tailoring \zsxlt to each target language and remedying for the varying performance of Max.\,\srcdev in \zsxlt to \textit{many} target languages.
Such a variation has been shown to be particularly pronounced in \zsxlt on token-level tasks like NER or POS \citep{schmidt2023free}.
On TyDiQA, we believe that accumulative averaging (slightly) better stabilizes the transfer from a small training set ($3.7$K instances).
%
\textsc{soup}s however perform notably worse than both \trgdev and accumulative averaging on NLI and NER. \textsc{soup}s lack the beneficial diversity of different runs, as the best snapshots often come from the same ``good'' run.\footnote{Extending \textsc{soup} to average the top-$10$ best snapshots does not improve performance.} 
Anecdotal evidence further exemplifies why source-language validation is inapt for \zsxlt. One of $63$ \srcdev models replicates XNLI results of \citet{conneau-etal-xlmr}, vastly exceeding all other runs (c.$\Delta{+}1.0$). This ``miraculous'' run though merely ranks 3rd according to source-language validation performance.

The above suggests that even the more sophisticated hyperparameter tuning strategies (e.g., Bayesian optimization) are unlikely to improve \zsxlt without target-language validation. On the other hand, accumulative averaging improves \zsxlt threefold: \textbf{(1)} Unlike model selection, it does not plateau in \zsxlt on suboptimal single runs that maximize source-language performance;
\textbf{(2)} \trgdev showcases that accumulative averaging ingests further runs with snapshots that perform well on \zsxlt;
\textbf{(3)} Model averaging irons out idiosyncratic noise of individual runs, leading to better performance. This renders accumulative averaging a robust (i.e., replicable results) and fair (i.e. true zero-shot) evaluation protocol for \zsxlt.    

\section{Conclusion}
\label{sec:conclusion}
Inconsistent hyperparameter tuning and model selection protocols exacerbate replicating previous results on \zsxlt.
In this focused study, we devise a \zsxlt evaluation protocol that addresses previous shortcomings and feeds two birds with one scone.
We show that accumulatively averaging snapshots -- rather than selecting models based on source-language validation performance -- both improves and stabilizes \zsxlt. Conventional model selection strategies prematurely settle for models that maximize source-language validation performance and discard runs that generalize better in \zsxlt. Accumulative model averaging both incorporates snapshots that transfer well and irons out models that perform badly. We find that model averaging correlates closely with ``oracle'' \zsxlt, which assumes models selection on target-language validation instances. We hope future work adopts model averaging to promote fair and reproducible \zsxlt that puts models on equal footing.

\section*{Limitations}

Additional factors must be taken into consideration, even though we aspire to evaluate \zsxlt on all levels of transparency (i.e., variants and strategies) across a varied set of downstream tasks on broad hyperparameter grids.
Neither model selection on source-language validation data nor accumulative averaging may benefit \zsxlt on certain tasks, as \citet{schmidt2023free}, e.g., do not find that any variant other than \trgdev yields gain over \last  on part-of-speech tagging. The underlying cause remains unclear. For instance, the gains on \zsxlt stemming from model selection or accumulative averaging likely depend on the type of distributional shift from the source-language training data and the target-language instances to transfer to (cf. \S\ref{sec:results-and-discussion}; e.g. dynamics of variants in \zsxlt for NER). accumulative averaging nevertheless remains a robust evaluation protocol, as \zsxlt performance is not expected to deteriorate via-\`a-vis other ``fair'' strategies (e.g., max.\,\srcdev).
In addition, there may exist a subset of pairs of learning rates and batch sizes that jointly maximize source- and target-language performance. 
However, as our results suggest (\S\ref{sec:results-and-discussion}), runs on such hyperparameters likely are indistinguishable from those that exclusively perform just as well on the source-language validation set.

\section*{Acknowledgments}
We thank the state of Baden-Württemberg for its support through access to the bwHPC. Ivan Vuli\'{c} is supported by a personal Royal Society University Research Fellowship \textit{`Inclusive and Sustainable Language Technology for a Truly Multilingual World'} (no 221137; 2022--).

\bibliography{anthology,custom}
\bibliographystyle{acl_natbib}

\appendix

\section{Appendix}
\label{sec:appendix}
\subsection{Reproduction Details}

\noindent \textbf{Code}. Our code is available at: \url{https://github.com/fdschmidt93/ofa-xlt}

\noindent \textbf{Model architectures.} All models use the \texttt{AutoModelFor\{SequenceClassification, TokenClassification, QuestionAnswering\}} of \texttt{xlm-roberta-large} for  the corresponding task from the \texttt{transformers} library \citep{wolf-etal-2020-transformers}.

\smallskip

\noindent \textbf{Compute Requirements.} We execute all experiments on a single V100 with 32GB VRAM.
We estimate that we require total compute time of c.1,050 hours over all fine-tuning iterations and evaluations.
We arrive at this budget as follows. We on average train models on NLI for about 11.5 hours, on TyDiQA-GoldP for roughly 1.5 hours, and on NER for an estimated 3 hours.
We therefore execute 63 training runs (21 hyperparameter configurations ran on for 3 seeds, cf. \S\ref{sec:experimental-setup}) for 16 hours for a total of c.1K GPU hours.
We loosely estimate that accumulative averaging adds another 50 hours of runtime for evaluation.

\smallskip

\noindent \textbf{Model Averaging.} We follow \citet{schmidt2023free} to enabling accumulative averaging of checkpoints for NLI and TyDiQA-GoldP. For these tasks, we initially fine-tune XLM-R$_{\text{large}}$ with a batch size of 32 and a learning rate of $2e^{{-}5}$. For NER, we find that merely randomly initializing the tasks heads across all runs with the same head slightly improves performance (~$\Delta+1.0$) of all variants in single-run and accumulative averaging. We suspect that the original language modelling weights better align with NER as a token-level classification task and do not diverge to incompatible sets of parameters in fine-tuning (cf. \S3 of \citet{schmidt2023free}).

\subsection{Full Results By Hyperparameter Configuration}
\label{appendix:full-results}

\begin{table*}[t!]
 \tiny
  \begin{adjustbox}{width=\linewidth,center}
   \setlength{\tabcolsep}{2.5pt}
    \begin{tabular}{cc|cccc|cccc|cccc}
    \multicolumn{14}{l}{\textbf{\zsxlt Performance}} \\
      \toprule
      \multicolumn{2}{c}{\textbf{Hyperparameters}} & \multicolumn{4}{c}{\textbf{NLI}} & \multicolumn{4}{c}{\textbf{TyDiQA-GoldP}} & \multicolumn{4}{c}{\textbf{NER}} \\
      \cmidrule(lr){1-2} \cmidrule(lr){3-6} \cmidrule(lr){7-10} \cmidrule(lr){11-14}
      Learning Rate & Batch Size & \textbf{\last} & \textbf{\srcdev} & \textbf{\ca} & \textbf{\trgdev} & \textbf{\last} & \textbf{\srcdev} & \textbf{\ca} & \textbf{\trgdev} & \textbf{\last} & \textbf{\srcdev} & \textbf{\ca} & \textbf{\trgdev} \\ \hline
                & $16$ & \cellcolor{green!45}{$77.2_{0.1}$         }  & \cellcolor{green!45}{$77.5_{0.3}$} & $77.6_{0.2}$                        & $77.9_{0.5}$                               & $71.3_{0.3}$                      & $71.3_{0.1}$                       & $68.4_{0.8}$                       & $71.4_{0.3}$                                & $\mathbf{45.9_{0.3}}$               & $\mathbf{45.9_{0.3}}$              & $\mathbf{46.5_{0.3}}$     & $47.8_{0.1}$ \\
$1e^{{-}6}$     & $32$ & \cellcolor{green!45}{$\mathbf{77.4_{0.1}}$}  & $77.5_{0.2}$                       & $77.7_{0.2}$                        & $78.1_{0.1}$                               & \cellcolor{green!45}{$71.2_{0.2}$} & \cellcolor{green!45}{$71.3_{0.4}$} & $63.2_{0.3}$                       & $71.6_{0.2}$                               & $45.3_{0.2}$                        & $45.3_{0.2}$                       & $44.7_{0.3}$              & $45.8_{0.3}$ \\
                & $64$ & $77.6_{0.1}$                                & $77.6_{0.3}$                       & $77.4_{0.1}$                        & $77.8_{0.1}$                                & $70.9_{0.4}$                      & $70.9_{0.4}$                       & $49.8_{0.2}$                       & $70.9_{0.1}$                                & $45.3_{0.2}$                        & $45.3_{0.2}$                       & $42.6_{0.3}$              & $45.5_{0.2}$ \\ \hline
                & $16$ & $76.7_{0.2}$                                & $76.9_{0.2}$                       & $77.5_{0.3}$                        & $77.7_{1.1}$                                & $71.6_{0.2}$                      & $71.9_{0.7}$                       & $72.2_{0.1}$                       & $72.0_{0.4}$                                & $44.0_{0.9}$                        & $44.9_{0.9}$                       & $47.4_{0.3}$              & \cellcolor{green!45}{$49.8_{0.3}$}\\
$5e^{{-}6}$     & $32$ & $76.8_{0.0}$                                & $77.6_{0.4}$                       & $77.6_{0.1}$                        & $78.3_{0.5}$                                & $71.6_{0.1}$                      & $71.5_{0.1}$                       & $71.6_{0.1}$                       & $72.3_{0.4}$                                & $42.8_{0.3}$                        & $42.9_{0.2}$                       & $45.8_{0.6}$              & $47.7_{1.0}$ \\
                & $64$ & $77.0_{0.2}$                                & $\mathbf{78.1_{0.6}}$              & $\mathbf{77.8_{0.2}}$               & \cellcolor{green!45}{$\mathbf{78.3_{0.3}}$} & $71.0_{0.9}$                      & $70.9_{0.6}$                       & $69.0_{0.7}$                       & $71.7_{0.1}$                                & $43.8_{1.6}$                        & $43.8_{1.6}$                       & $46.2_{1.3}$              & $49.0_{0.4}$ \\ \hline
                & $16$ & $76.6_{0.2}$                                & $76.6_{0.2}$                       & $77.5_{0.2}$                        & $77.1_{0.2}$                                & $73.0_{0.4}$                      & $72.5_{0.4}$                       & $73.5_{0.5}$                       & $73.7_{0.6}$                                & $40.6_{0.1}$                        & \cellcolor{green!45}{$40.7_{1.8}$} & \cellcolor{green!45}{$43.9_{1.4}$} & $48.0_{2.5}$ \\ 
$1e^{{-5}}$     & $32$ & $76.8_{0.2}$                                & $77.1_{0.4}$                       & $77.6_{0.1}$                        & $77.2_{0.2}$                                & $72.1_{0.4}$                      & $72.6_{0.5}$                       & $73.0_{0.3}$                       & $73.2_{0.3}$                                & $40.1_{1.8}$                        & $40.1_{1.8}$                       & $42.8_{1.5}$              & $46.0_{3.1}$ \\
                & $64$ & $76.8_{0.3}$                                & $77.2_{0.5}$                       & \cellcolor{green!45}{$77.5_{0.2}$}  & $77.7_{0.2}$                                & $72.0_{0.8}$                      & $71.7_{0.8}$                       & $71.7_{0.2}$                       & $73.0_{0.3}$                                & $43.1_{1.9}$                        & $43.4_{2.3}$                       & $46.6_{1.2}$              & $49.7_{0.7}$ \\ \hline
                & $16$ & $75.6_{0.1}$                                & $75.6_{0.2}$                       & $76.8_{0.3}$                        & $76.5_{0.4}$                                & $73.7_{0.5}$                      & $\mathbf{73.3_{0.3}}$              & $74.4_{0.6}$                       & \cellcolor{green!45}{$\mathbf{74.1_{0.4}}$} & \cellcolor{green!45}{$41.0_{3.1}$}  & $42.0_{2.6}$                       & $45.3_{2.0}$              & $\mathbf{50.0_{1.1}}$ \\
$1.5e^{{-5}}$   & $32$ & $76.6_{0.1}$                                & $76.5_{0.1}$                       & $77.4_{0.2}$                        & $77.0_{0.1}$                                & $73.0_{0.7}$                      & $73.1_{0.9}$                       & $74.0_{0.1}$                       & $73.7_{0.6}$                                & $39.9_{1.1}$                        & $40.6_{1.3}$                       & $43.3_{0.2}$              & $46.2_{1.9}$ \\
                & $64$ & $76.8_{0.1}$                                & $77.1_{0.5}$                       & $77.6_{0.3}$                        & $77.7_{0.6}$                                & $72.9_{0.5}$                      & $72.6_{0.7}$                       & $73.3_{0.4}$                       & $73.5_{0.3}$                                & $40.6_{2.1}$                        & $41.1_{2.0}$                       & $42.8_{1.5}$              & $46.0_{1.1}$ \\ \hline
                & $16$ & $74.1_{0.3}$                                & $74.1_{0.3}$                       & $76.1_{0.2}$                        & $74.8_{0.2}$                                & $72.9_{0.5}$                      & $72.7_{0.2}$                       & $74.1_{0.2}$                       & $73.5_{0.2}$                                & $38.6_{1.2}$                        & $38.6_{1.2}$                       & $43.8_{1.4}$              & $46.2_{3.0}$ \\
$2e^{{-5}}$     & $32$ & $75.9_{0.3}$                                & $76.1_{0.4}$                       & $77.1_{0.1}$                        & $76.4_{0.5}$                                & $73.1_{0.1}$                      & $72.3_{0.9}$                       & \cellcolor{green!45}{$74.1_{0.2}$} & $73.3_{0.4}$                                & $39.3_{0.7}$                        & $39.0_{1.1}$                       & $42.3_{1.5}$              & $44.3_{2.5}$ \\
                & $64$ & $76.6_{0.5}$                                & $77.0_{0.4}$                       & $77.4_{0.1}$                        & $77.7_{0.2}$                                & $71.9_{0.4}$                      & $71.7_{1.1}$                       & $73.3_{0.5}$                       & $72.8_{0.4}$                                & $39.5_{1.2}$                        & $39.7_{1.3}$                       & $42.3_{1.1}$              & $46.2_{0.4}$ \\ \hline
                & $16$ & $71.5_{0.2}$                                & $71.3_{0.4}$                       & $75.0_{0.2}$                        & $73.1_{0.1}$                                & $\mathbf{73.2_{0.4}}$             & $72.4_{0.9}$                       & $74.7_{0.1}$                       & $73.4_{0.5}$                                & $39.3_{1.0}$                        & $39.3_{1.0}$                       & $44.3_{1.7}$              & $47.1_{0.6}$ \\
$2.5e^{{-5}}$   & $32$ & $74.8_{0.2}$                                & $74.8_{0.2}$                       & $76.3_{0.1}$                        & $76.1_{0.6}$                                & $72.3_{0.2}$                      & $71.7_{1.0}$                       & $74.8_{0.3}$                       & $73.4_{0.3}$                                & $39.4_{1.5}$                        & $39.4_{1.5}$                       & $42.5_{0.6}$              & $44.5_{0.7}$ \\
                & $64$ & $76.7_{0.2}$                                & $76.7_{0.2}$                       & $77.5_{0.1}$                        & $77.3_{0.5}$                                & $72.6_{0.6}$                      & $72.3_{0.7}$                       & $74.2_{0.1}$                       & $73.3_{1.0}$                                & $39.9_{1.0}$                        & $39.9_{1.0}$                       & $43.5_{1.5}$              & $47.7_{2.3}$ \\ \hline
                & $16$ & $67.9_{0.6}$                                & $67.7_{0.2}$                       & $73.0_{0.3}$                        & $70.8_{0.9}$                                & $72.3_{0.2}$                      & $71.7_{0.4}$                       & $74.4_{0.4}$                       & $72.4_{0.5}$                                & $37.4_{1.5}$                        & $37.3_{1.1}$                       & $43.0_{0.8}$              & $46.3_{2.5}$ \\
$3e^{{-5}}$     & $32$ & $73.3_{0.1}$                                & $73.3_{0.4}$                       & $75.8_{0.2}$                        & $74.4_{0.1}$                                & $71.7_{0.4}$                      & $71.6_{0.7}$                       & $\mathbf{75.1_{0.1}}$              & $73.6_{0.8}$                                & $37.9_{1.6}$                        & $38.0_{1.8}$                       & $43.7_{1.6}$              & $47.2_{2.7}$ \\
                & $64$ & $75.6_{0.1}$                                & $75.4_{0.3}$                       & $76.8_{0.3}$                        & $76.3_{0.7}$                                & $72.0_{0.2}$                      & $71.8_{0.4}$                       & $74.1_{0.5}$                       & $73.2_{0.4}$                                & $39.0_{1.9}$                        & $39.4_{1.4}$                       & $42.1_{1.2}$              & $44.0_{1.2}$ \\ \hline
& $\mathbf{\Delta}$ & $0.0$ & $0.6$ & $0.3$ & $0.0$ & $2.0$ & $2.0$ & $1.0$ & $0.0$ & $4.9$ & $5.2$ & $2.6$ & $0.2$ \\\hline
      \bottomrule
    \end{tabular}
  \end{adjustbox}
  \caption{\zsxlt averaged over all target languages by task, model variant, and hyperparameters (cf. \S\ref{sec:experimental-setup}). For each column, \textbf{best \zsxlt emphasized in bold} and \colorbox{green!45}{max. validation performance (cf. Table \ref{tab:xlt-full-validation-performance}) shaded in green}. $\mathbf{\Delta}$ is the difference of \textbf{best \zsxlt} and \colorbox{green!45}{\zsxlt on models that maximize validation performance.}  \textbf{Metrics:} accuracy for NLI, span-$F_1$ for TyDiQA and token-level $F_1$ for NER. Subscripts denote std. deviation. }
  \label{tab:xlt-full-results}
  \vspace{-0.5em}
\end{table*}

\begin{table*}[t!]
 \tiny
  \begin{adjustbox}{width=\linewidth,center}
   \setlength{\tabcolsep}{2.5pt}
    \begin{tabular}{cc|cccc|cccc|cccc}
    \multicolumn{14}{l}{\textbf{Validation Set Performance}} \\
      \toprule
      \multicolumn{2}{c}{\textbf{Hyperparameters}} & \multicolumn{4}{c}{\textbf{NLI}} & \multicolumn{4}{c}{\textbf{TyDiQA-GoldP}} & \multicolumn{4}{c}{\textbf{NER}} \\
      \cmidrule(lr){1-2} \cmidrule(lr){3-6} \cmidrule(lr){7-10} \cmidrule(lr){11-14}
      Learning Rate & Batch Size & \textbf{\last} & \textbf{\srcdev} & \textbf{\ca} & \textbf{\trgdev} & \textbf{\last} & \textbf{\srcdev} & \textbf{\ca} & \textbf{\trgdev} & \textbf{\last} & \textbf{\srcdev} & \textbf{\ca} & \textbf{\trgdev} \\ \hline
                & $16$ & $\mathbf{90.2_{0.2}}$  & $\mathbf{90.5_{0.2}}$     & $90.1_{0.2}$          & $79.2_{0.3}$          & $76.4_{0.2}$          & $76.7_{0.5}$          & $75.1_{0.3}$          & $67.0_{0.2}$          & $81.9_{0.1}$          & $81.9_{0.1}$          & $79.5_{0.2}$          & $49.9_{0.1}$          \\
$1e^{{-}6}$     & $32$ & $\mathbf{90.2_{0.2}}$  & $90.3_{0.1}$              & $90.0_{0.0}$          & $79.2_{0.1}$          & $\mathbf{77.0_{0.5}}$ & $\mathbf{77.9_{0.8}}$ & $73.5_{0.4}$          & $66.9_{0.1}$          & $80.4_{0.1}$          & $80.4_{0.1}$          & $76.5_{0.3}$          & $48.3_{0.3}$          \\
                & $64$ & $90.1_{0.1}$           & $90.2_{0.1}$              & $89.6_{0.1}$          & $78.9_{0.0}$          & $76.0_{0.6}$          & $76.3_{0.4}$          & $64.4_{0.2}$          & $66.5_{0.2}$          & $78.0_{0.1}$          & $78.0_{0.1}$          & $71.4_{0.4}$          & $47.7_{0.2}$          \\ \hline
                & $16$ & $89.3_{0.4}$           & $89.7_{0.2}$              & $90.1_{0.2}$          & $78.9_{0.6}$          & $73.8_{0.9}$          & $76.0_{0.5}$          & $75.9_{0.8}$          & $68.7_{0.1}$          & $85.1_{0.3}$          & $85.3_{0.4}$          & $84.8_{0.1}$          & $\mathbf{52.1_{0.4}}$ \\
$5e^{{-}6}$     & $32$ & $89.5_{0.4}$           & $89.9_{0.2}$              & $90.1_{0.2}$          & $79.4_{0.4}$          & $74.5_{0.5}$          & $75.8_{0.7}$          & $76.2_{0.8}$          & $68.2_{0.2}$          & $84.7_{0.1}$          & $84.7_{0.1}$          & $84.0_{0.2}$          & $50.0_{1.2}$          \\
                & $64$ & $89.6_{0.1}$           & $90.2_{0.3}$              & $90.0_{0.1}$          & $\mathbf{79.4_{0.3}}$ & $75.1_{0.7}$          & $76.2_{0.5}$          & $75.5_{0.4}$          & $67.8_{0.2}$          & $84.2_{0.2}$          & $84.2_{0.2}$          & $83.0_{0.3}$          & $51.2_{0.4}$          \\ \hline
                & $16$ & $88.9_{0.3}$           & $89.1_{0.2}$              & $89.6_{0.1}$          & $78.2_{0.1}$          & $74.5_{0.4}$          & $75.7_{0.1}$          & $75.8_{0.6}$          & $69.4_{0.3}$          & $85.6_{0.1}$          & $\mathbf{85.7_{0.1}}$ & $\mathbf{85.8_{0.2}}$ & $50.3_{2.2}$          \\ 
$1e^{{-5}}$     & $32$ & $89.1_{0.1}$           & $89.3_{0.2}$              & $89.6_{0.1}$          & $78.6_{0.3}$          & $74.4_{0.6}$          & $75.9_{0.6}$          & $75.7_{1.0}$          & $69.2_{0.1}$          & $85.4_{0.1}$          & $85.5_{0.2}$          & $85.3_{0.2}$          & $48.4_{3.1}$          \\
                & $64$ & $89.7_{0.4}$           & $89.9_{0.0}$              & $\mathbf{90.1_{0.1}}$ & $79.1_{0.3}$          & $74.8_{1.2}$          & $76.1_{0.0}$          & $76.7_{0.4}$          & $69.0_{0.2}$          & $85.0_{0.1}$          & $85.0_{0.1}$          & $84.5_{0.1}$          & $52.0_{0.8}$          \\ \hline
                & $16$ & $88.4_{0.3}$           & $88.5_{0.2}$              & $89.1_{0.2}$          & $77.5_{0.2}$          & $74.8_{0.7}$          & $76.3_{0.3}$          & $76.5_{0.7}$          & $\mathbf{69.7_{0.2}}$ & $\mathbf{86.0_{0.1}}$ & $86.1_{0.1}$          & $86.2_{0.1}$          & $52.0_{1.3}$          \\
$1.5e^{{-5}}$   & $32$ & $89.0_{0.5}$           & $89.0_{0.5}$              & $89.6_{0.1}$          & $78.2_{0.2}$          & $75.1_{0.7}$          & $76.5_{0.4}$          & $76.5_{0.3}$          & $69.5_{0.4}$          & $85.4_{0.2}$          & $85.5_{0.2}$          & $85.8_{0.1}$          & $48.3_{1.6}$          \\
                & $64$ & $89.0_{0.3}$           & $89.4_{0.3}$              & $89.5_{0.2}$          & $78.7_{0.3}$          & $75.5_{0.9}$          & $76.2_{0.3}$          & $76.3_{0.8}$          & $69.3_{0.2}$          & $85.1_{0.2}$          & $85.2_{0.1}$          & $85.2_{0.3}$          & $48.3_{1.1}$          \\ \hline
                & $16$ & $87.7_{0.6}$           & $87.9_{0.3}$              & $88.9_{0.4}$          & $75.8_{0.2}$          & $74.6_{0.7}$          & $76.5_{0.8}$          & $76.8_{0.7}$          & $69.3_{0.1}$          & $85.9_{0.2}$          & $85.9_{0.2}$          & $86.0_{0.2}$          & $48.2_{3.1}$          \\
$2e^{{-5}}$     & $32$ & $88.7_{0.2}$           & $88.8_{0.1}$              & $89.3_{0.4}$          & $77.6_{0.3}$          & $76.5_{1.5}$          & $77.1_{1.0}$          & $\mathbf{78.1_{0.5}}$ & $69.4_{0.5}$          & $85.4_{0.2}$          & $85.4_{0.2}$          & $85.7_{0.2}$          & $46.8_{2.5}$          \\
                & $64$ & $89.3_{0.2}$           & $89.4_{0.1}$              & $89.8_{0.1}$          & $78.7_{0.2}$          & $73.9_{0.7}$          & $76.3_{0.3}$          & $76.9_{0.5}$          & $69.2_{0.1}$          & $85.2_{0.2}$          & $85.4_{0.3}$          & $85.6_{0.3}$          & $48.4_{0.6}$          \\ \hline
                & $16$ & $87.5_{0.3}$           & $87.7_{0.1}$              & $88.6_{0.3}$          & $74.0_{0.3}$          & $75.8_{0.9}$          & $76.4_{0.2}$          & $77.2_{0.5}$          & $69.3_{0.3}$          & $85.5_{0.2}$          & $85.5_{0.2}$          & $86.0_{0.3}$          & $49.5_{0.5}$          \\
$2.5e^{{-5}}$   & $32$ & $88.6_{0.1}$           & $88.6_{0.1}$              & $89.0_{0.2}$          & $76.9_{0.4}$          & $74.7_{0.6}$          & $76.0_{0.3}$          & $77.0_{0.6}$          & $69.1_{0.1}$          & $85.7_{0.2}$          & $85.7_{0.2}$          & $86.0_{0.1}$          & $46.9_{0.6}$          \\
                & $64$ & $88.8_{0.3}$           & $88.9_{0.2}$              & $89.4_{0.2}$          & $78.4_{0.4}$          & $75.1_{1.8}$          & $76.1_{0.7}$          & $76.9_{0.8}$          & $69.2_{0.4}$          & $85.4_{0.0}$          & $85.4_{0.0}$          & $85.7_{0.3}$          & $49.8_{2.3}$          \\ \hline
                & $16$ & $86.7_{0.2}$           & $86.8_{0.2}$              & $87.7_{0.1}$          & $71.7_{0.7}$          & $74.1_{0.6}$          & $75.6_{1.5}$          & $76.1_{0.8}$          & $68.5_{0.2}$          & $85.5_{0.1}$          & $85.6_{0.1}$          & $86.2_{0.0}$          & $48.5_{2.4}$          \\
$3e^{{-5}}$     & $32$ & $87.8_{0.3}$           & $88.0_{0.5}$              & $89.0_{0.3}$          & $75.3_{0.3}$          & $73.8_{1.3}$          & $76.2_{1.0}$          & $76.5_{0.8}$          & $69.1_{0.2}$          & $85.4_{0.1}$          & $85.4_{0.1}$          & $86.0_{0.1}$          & $49.4_{3.0}$          \\
                & $64$ & $88.4_{0.2}$           & $88.5_{0.1}$              & $89.4_{0.2}$          & $77.4_{0.7}$          & $74.8_{0.6}$          & $76.3_{0.2}$          & $77.7_{0.4}$          & $69.1_{0.2}$          & $85.4_{0.1}$          & $85.4_{0.1}$          & $85.8_{0.1}$          & $46.2_{1.5}$          \\
      
      \bottomrule
    \end{tabular}
  \end{adjustbox}
  \caption{Validation performance by task, model variant, and hyperparameters (cf. \S\ref{sec:experimental-setup}). \last, \srcdev, and \ca validate on source-language validation splits; \trgdev denotes performance averaged over individual snapshots of a run that perform best by target-language validation set.
  For each column, \textbf{best validation performance in bold}. 
  \textbf{Metrics:} accuracy for NLI, span-$F_1$ for TyDiQA and token-level $F_1$ for NER. Subscripts denote std. deviation. 
  }
  \label{tab:xlt-full-validation-performance}
  \vspace{-0.5em}
\end{table*}

\end{document}